\documentclass[10pt]{article}

\usepackage[preprint]{tmlr}

\usepackage{amsmath,amssymb}
\usepackage{graphicx}
\usepackage{booktabs}
\usepackage{enumitem}
\usepackage{flafter}
\usepackage{placeins}
\usepackage{hyperref}
\usepackage{url}
\usepackage[nameinlink,capitalise,noabbrev]{cleveref}

\graphicspath{{figures/}}

\title{From Interaction Traces to Persistent Skills:\\
Online Evolution for Computer-Use Agents}
\author{\name Longtao Hu \email 2023090902005@std.uestc.edu.cn \\
      \addr University of Electronic Science and Technology of China
      \AND
      \name Xiao Liang \email 0314liangxiao@gmail.com \\
      \addr Zhejiang University
      \AND
      \name Linchao Zhu\thanks{Corresponding author.} \email zhulinchao7@gmail.com \\
      \addr Zhejiang University}

\hypersetup{
    hidelinks,
    pdftitle={From Interaction Traces to Persistent Skills: Online Evolution for Computer-Use Agents},
    pdfauthor={Longtao Hu, Xiao Liang, Linchao Zhu}
}

\begin{document}

\maketitle

\begin{abstract}
Computer-use agents can execute increasingly complex tasks in graphical
interfaces, but their interaction experience is typically transient:
procedural knowledge acquired from one rollout is not systematically retained,
refined, and reused in later tasks. Existing skill libraries provide external
procedural knowledge, yet their incremental value over the same agent operating
without skills---and their longitudinal dynamics under repeated
interaction---remain insufficiently characterized. We present an online
skill-evolution framework that converts interaction trajectories and evaluator
feedback into a persistent, versioned library of reusable procedures. Each
iteration executes against a frozen library snapshot, and evidence-guided skill
updates become available in subsequent iterations without changing model
parameters. We compare the full evolving-library system with a
configuration-matched empty-library control across four OSWorld application
domains under the same fixed action-generation and GUI-grounding stack, task
sets, and iteration horizons. Following a five-iteration empty-library warm-up,
Full attains a higher post-warm-up mean evaluator score in all four observed
domain runs, with mean differences of 5.7--18.6 percentage points and
domain-dependent temporal stability. In GIMP, provenance-aware analysis reveals
retrieval across task-of-origin boundaries and revision churn, where repeated
accepted edits fail to recover the originating task. These findings characterize
evolving skill libraries as auditable, shared procedural memory that can improve
a fixed computer-use stack, while showing that their benefits are conditional
and repeated revision does not guarantee recovery. Code is released at
\url{https://github.com/LongtaoHu/Skill-Evo4GUI.git}.
\end{abstract}


\section{Introduction}
\label{sec:introduction}

Computer-use agents translate natural-language instructions and screen
observations into keyboard and mouse actions, allowing them to operate graphical
software without application-specific APIs. WebArena and OSWorld provide
reproducible evaluation of long-horizon interaction in functional websites and
desktop applications \citep{zhou2023webarena,xie2024osworld}. Unless explicitly
externalized, however, this experience remains transient: after a rollout,
successful procedures, failure evidence, and application-specific operational
knowledge are not systematically retained. Consequently, agents may rediscover
the same procedure or reproduce the same failure across interactions.

\begin{figure}[t]
    \centering
    \includegraphics[width=\linewidth]{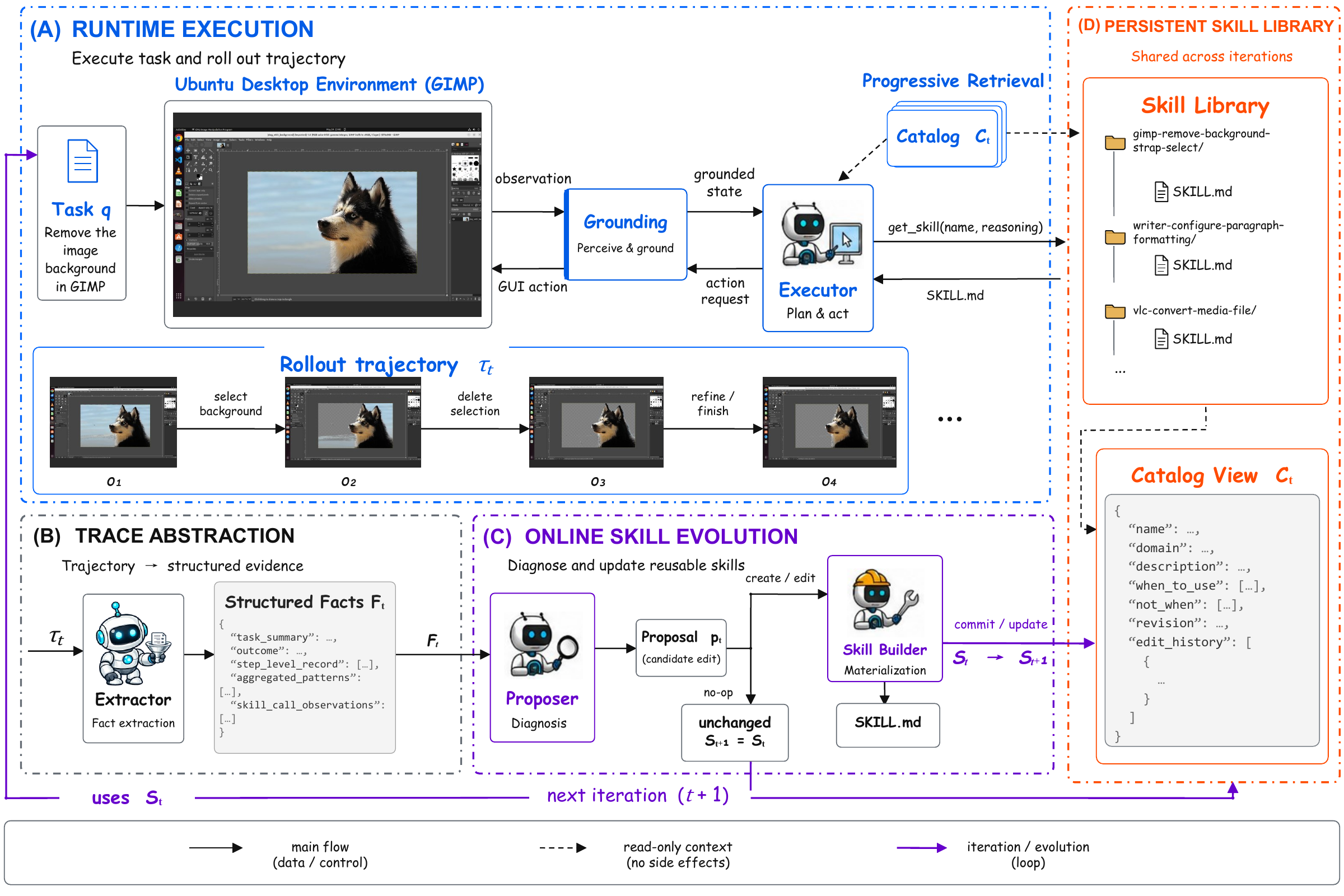}
    \caption{Online skill evolution. Execution under $S_t$ produces trajectories
    that are abstracted and diagnosed to create, edit, delete, or preserve
    skills. Accepted changes form $S_{t+1}$. The Executor sees only a lightweight
    catalog unless it explicitly retrieves a complete skill body.}
    \label{fig:framework}
\end{figure}
\FloatBarrier

External memory offers a way to retain and reuse interaction experience without
updating model parameters. Prior agents store verbal reflections or distilled
experiential insights \citep{shinn2023reflexion,zhao2023expel}, while subsequent
systems represent reusable behavior as workflows, executable programs, or
structured skills
\citep{wang2023voyager,wang2024awm,chen2026cuaskill,zhang2026mmskills}. Recent
evaluations show that skills can improve performance, but successful use depends
on retrieving and correctly following the relevant procedure
\citep{li2026skillsbench,han2026skilluse}. Yet two issues remain insufficiently
characterized for skill libraries that evolve online in grounded desktop
agents. First, does an evolving library provide incremental value over repeated
interaction relative to the same computer-use system operating without skills?
Second, does the resulting library support reuse beyond the tasks that produced
its skills, and how do individual skills evolve under repeated success and
failure? A final library or aggregate endpoint score cannot reveal these
longitudinal dynamics or the provenance of the procedures involved.

We study these questions through an online skill-evolution framework for
grounded computer use. The action-generation and GUI-grounding models remain
fixed, so adaptation occurs only through a persistent external skill library.
As illustrated in \cref{fig:framework}, every task within an iteration executes
against the same frozen snapshot. Trajectories and evaluator feedback are then
converted into structured evidence that guides skill creation, revision, or
preservation. Accepted updates are versioned but become visible only through
the next iteration snapshot, separating execution from evolution. At runtime,
a lightweight catalog exposes skill metadata, while complete procedures are
retrieved on demand. The same records preserve the provenance needed to trace
how skills are created, reused, and revised. This design makes adaptation
observable without conflating skill evolution with model fine-tuning or
within-rollout adaptation.

We evaluate the integrated framework through a configuration-matched
longitudinal comparison across four OSWorld application domains. The Full
condition evolves and retrieves skills after a five-iteration empty-library
warm-up, whereas the empty-library control uses the same fixed action-generation
and GUI-grounding stack, task sets, and iteration horizons while keeping the
library empty throughout. In the observed runs, Full achieves higher
post-warm-up mean evaluator score in every domain, with mean differences relative to the
control ranging from 5.7 to 18.6 percentage points. The separation is neither
uniform nor monotonic: its magnitude and temporal stability vary substantially
across domains. In GIMP, provenance-aware analysis further identifies retrieval
across task-of-origin boundaries, providing direct evidence that part of the
learned library is reused as shared procedural memory rather than retained only
as task-specific records. The same analysis also exposes revision churn, in
which continued failures trigger repeated accepted edits without producing task
recovery. Together, the configuration-matched longitudinal comparison and
provenance analysis show that an evolving skill library can add measurable value
to a fixed computer-use stack, while its benefits remain domain-dependent and
repeated revision does not guarantee improvement.

\paragraph{Contributions.} Our contributions are threefold:

\begin{itemize}[leftmargin=*]
    \item We present an online skill-evolution framework that converts
    interaction trajectories and evaluator feedback into a persistent,
    versioned library of reusable procedures. By separating iteration-frozen
    execution from evidence-guided skill updates, the framework enables
    continual adaptation without modifying the underlying Executor or Grounding
    models.

    \item We demonstrate the value of the integrated evolving-library system
    through a matched longitudinal comparison across four OSWorld application
    domains. Under the same fixed Executor--Grounding stack, task sets, and
    iteration horizons, Full achieves higher post-warm-up mean evaluator score than the
    empty-library control in every domain, with meaningful differences in the
    magnitude and temporal stability of the performance advantage.

    \item We introduce a provenance-aware analysis that links each skill to its
    originating task, subsequent retrievals, revision history, and downstream
    execution outcomes. Within the analyzed GIMP skill histories, the analysis
    demonstrates retrieval across task-of-origin boundaries rather than only
    task-specific reuse.
\end{itemize}

\section{Related Work}
\label{sec:related-work}

\paragraph{Interactive environments and execution foundations.}
WebArena brought reproducible, long-horizon evaluation to functional websites
\citep{zhou2023webarena}, while OSWorld extended executable evaluation to
open-ended tasks in real desktop applications \citep{xie2024osworld}. Recent
systems improve this substrate from complementary directions. Cradle studies a
unified screenshot-in, keyboard-and-mouse-out interface across games and
software \citep{tan2024cradle}, while EvoCUA learns a native computer-use policy
from large-scale synthetic experience
\citep{xue2026evocua}; Qwen3 and Kimi K2.5 represent increasingly capable
general and visual-agentic backbones \citep{yang2025qwen3,kimi2026k25}.
SeeClick isolates GUI grounding as a core visual-agent challenge
\citep{cheng2024seeclick}, and MVP improves coordinate prediction through
training-free multi-view inference \citep{zhang2025mvp}. These advances
strengthen models, data, or grounding. Our question is orthogonal: with the
Executor and Grounding models fixed, what
procedural knowledge can be accumulated externally through repeated desktop
interaction?

\paragraph{Reusable procedural knowledge.}
Non-parametric improvement first retained feedback and experience as external
language memory: Reflexion stores verbal self-feedback \citep{shinn2023reflexion},
whereas ExpeL distills reusable insights from collections of task experience
\citep{zhao2023expel}. Voyager then stores executable code skills for open-ended
embodied exploration
\citep{wang2023voyager}, and Agent Workflow Memory induces reusable routines
from offline or online web trajectories \citep{wang2024awm}. More recent work
specializes skill representations for computer use. CUA-Skill provides an
engineered Windows skill base with parameterized execution and composition
graphs \citep{chen2026cuaskill}; MMSkills combines textual procedures, runtime
state cards, and visual keyframes \citep{zhang2026mmskills}; and MMG2Skill
distills in-the-wild guides into editable skills refined by trajectory feedback
\citep{che2026mmg2skill}. These systems establish that procedures can be
represented, retrieved, and composed beyond a raw trajectory. We likewise use
progressive retrieval and focus on how a persistent library changes across
repeated application-level rollouts.

\paragraph{Self-improving skill libraries.}
Several frameworks turn skill maintenance into an explicit learning process.
SAGE couples sequential skill accumulation with reinforcement learning
\citep{wang2025sage}; EvoSkill proposes creations and edits from failure
analysis and retains changes through held-out validation \citep{alzubi2026evoskill};
and SkillOS trains a curator to update a repository used by a frozen executor
\citep{ouyang2026skillos}. XSkill continually links visually grounded experience
and skill memories \citep{jiang2026xskill}, while Mem2Evolve co-evolves
experience and dynamically created assets \citep{cheng2026mem2evolve}.
MUSE-Autoskill further unifies creation, reuse, management, evaluation, and
refinement in a skill lifecycle \citep{lin2026museautoskill}. Complementary to
these general frameworks, we foreground a longitudinal GUI-system view:
iteration-frozen execution, evidence-constrained and versioned mutations, and
task-level histories that expose both improvement and failure-driven
over-editing.

\paragraph{Evaluating skill use.}
SkillsBench measures skill efficacy through paired executions with and without
curated skill packages \citep{li2026skillsbench}. Skill-Use instead isolates
whether an agent triggers a relevant skill under progressive disclosure,
complies with its procedure, and respects its boundaries \citep{han2026skilluse}.
These findings caution that possessing a skill is not equivalent to using it
reliably. Accordingly, our study records retrieval coverage and associates
calls with subsequent outcomes, while focusing on the evolution of one
persistent library per application domain rather than proposing another broad
skill benchmark.

\section{Method}
\label{sec:method}

We study how a GUI agent can revise persistent procedural knowledge from its
own experience while its Executor and Grounding models remain fixed. As shown in
\cref{fig:framework}, the pipeline separates read-only execution under a frozen
skill state $S_t$ from write-enabled evolution. Each iteration executes a fixed
task set, abstracts the resulting trajectories, and selectively materializes
skill changes to form $S_{t+1}$.

\subsection{Iterative Setting and Library State}
\label{sec:iterative-setting}

For application domain $d$, let
$\mathcal{Q}_d=\{q_1,\ldots,q_{N_d}\}$ denote a fixed task set and let one
iteration be a complete sweep over this set. We study two independently
executed longitudinal conditions,
$c\in\{\mathrm{Full},\mathrm{Empty}\}$, using the same task order. At the start
of iteration $t$, the current live library is exported as a read-only snapshot
$S_t^c$ shared by all tasks in that sweep. Each trajectory is therefore
generated under a single iteration-frozen skill state:
\begin{equation}
    \tau_{t,i}^{c}=\operatorname{Execute}(q_i;S_t^{c}).
    \label{eq:iterative-execution}
\end{equation}

Both conditions begin with an empty library, but only Full may accumulate
persistent skills:
\begin{equation}
    S_t^{\mathrm{Full}}=\varnothing\quad(t=0,\ldots,4),
    \qquad
    S_t^{\mathrm{Empty}}=\varnothing\quad(\forall t).
    \label{eq:library-conditions}
\end{equation}
After the Full sweep at iteration $t$, trajectory evidence is processed and
accepted mutations may update the writable live library. These mutations are
never exposed to execution within the current sweep; they first become
available when the next snapshot $S_{t+1}^{\mathrm{Full}}$ is exported. The
first five task-specific observations are thus collected under empty snapshots,
and mutations accepted after iteration 4 form the first non-empty execution
snapshot, $S_5^{\mathrm{Full}}$.

Empty retains the same runtime skill interface, including the catalog prompt
and \texttt{get\_skill}, but the catalog remains empty and no skill can be
returned. It runs the same Extractor while disabling the Proposer and Skill
Builder, thereby clamping the library state to $\varnothing$. Because Full and
Empty are separate longitudinal runs rather than a shared warm-up, their
observed trajectories may differ even during iterations 0--4. Here,
\emph{online evolution} denotes persistent iteration-level updates, not
within-rollout adaptation or parameter training.

\subsection{Runtime Execution and Progressive Retrieval}
\label{sec:runtime-retrieval}

The runtime path keeps procedural guidance separate from GUI actuation. Given a
task instruction and the current screenshot, the fixed Executor selects a tool
call. For coordinate-dependent actions, it describes the target interface
element in natural language; the fixed Grounding model maps this description
and screenshot to an absolute pixel coordinate, which is executed through
\texttt{pyautogui}. The resulting screenshot becomes the next observation.
Only executable GUI action calls enter $\tau_{t,i}^{c}$, while non-GUI calls
such as skill retrieval are recorded separately in telemetry. Skills do not
execute GUI actions directly: skill-conditioned decisions still pass through
the same Executor tool interface, and coordinate-dependent actions continue to
require the fixed Grounding model.

Progressive retrieval exposes stored procedures without placing the entire
library in the Executor context. At the start of each iteration, the system
parses only the \texttt{name} and \texttt{description} fields from each
\texttt{SKILL.md} in the frozen snapshot and dynamically renders them as a
Markdown catalog in the Executor prompt; no separate catalog file or database
is maintained. When a catalog entry appears relevant, the Executor calls
\texttt{get\_skill(name, reasoning)} to retrieve its complete procedural body
as a tool result, with \texttt{reasoning} recording the retrieval rationale.
Both catalog access and \texttt{get\_skill} are read-only, so retrieval cannot
mutate the live library or expose revisions committed during the current
iteration. Under Empty, the same interface renders an empty catalog and returns
no skill body.

\subsection{Trace Abstraction and Provenance}
\label{sec:trace-abstraction}

Raw interaction traces are too detailed for consistent cross-iteration
diagnosis, so the Extractor converts each trajectory $\tau_{t,i}^{c}$ and its
auxiliary evidence $E_{t,i}^{c}$ into structured facts $F_{t,i}^{c}$. Inputs
combine executed actions and selected post-action screenshots with the official
evaluator score, postcondition checks, and telemetry that authoritatively
records each skill call's occurrence, identity, iteration, and returned
revision. The facts summarize task outcome, step-level state--action--effect
observations, recurring patterns, capability limitations, and skill-call
observations. The recorded evaluator score overrides model restatements. Skill
calls mentioned by the Extractor but absent from telemetry are removed;
telemetry-recorded calls omitted by the Extractor are restored with an
indeterminate outcome association. In Full, these facts provide the Proposer
with a compact cross-iteration history; under Empty, the same extraction stage
runs without triggering library evolution.

Together, structured facts, skill-call telemetry, and versioned library
metadata preserve provenance across skill creation, revision, and retrieval. A
skill's origin task is fixed when it is created; every accepted edit separately
records the contributing task and resulting revision. Each retrieval records
the consumer task, iteration, delivered revision, and an observational
association with the downstream rollout---task success, continued failure,
partial progress, or an indeterminate relation. We call a retrieval cross-origin
when its consumer differs from the skill's origin, even if other tasks later
revise that skill. Telemetry establishes what was retrieved, not its effect on
execution. Accordingly, these associations support tracing reuse and revision
histories but not causal attribution: one rollout may retrieve several skills,
and its outcome may depend on other actions, model behavior, or environment
state.

\subsection{Evidence-Guided Skill Evolution and Versioned Persistence}
\label{sec:skill-evolution}

In Full, the Proposer diagnoses an up-to-five-iteration structured-fact history
for the same task together with current live-library metadata. It emits a
proposal from \texttt{create\_new}, \texttt{edit\_existing},
\texttt{delete\_existing}, \texttt{no\_op}, or \texttt{unresolved}, with a
target and rationale where required. Create and edit proposals specify a
high-level intent rather than a complete skill document; \texttt{no\_op}
preserves the library, while \texttt{unresolved} records that the available
evidence does not support a library action.

A Coordinator converts each proposal into a validated final action before any
mutation occurs. It enforces the decision schema and domain-specific skill-name
prefix. Creation requires five observations; duplicate creation names and edits
exceeding the size gate are degraded to \texttt{no\_op}; and
\texttt{unresolved} is accepted only when the evidence window contains at least
two failures. The original proposal and any degradation reason remain in
telemetry. For an accepted create or edit, the Skill Builder receives the
validated action's target name and high-level intent; an edit additionally
receives the targeted skill's current \texttt{SKILL.md}. The Builder generates
a complete skill document---for an edit, a full replacement rather than a
direct textual application of the Proposer's intent. An accepted deletion
removes the targeted skill without invoking the Builder. Each successful
mutation is written to the live library as an individual Git commit, preserving
its revision history and contributing task.

All reported experiments use serial decision mode. After the rollout sweep,
tasks enter the evolution stage in fixed task-ID order; the Proposer decision,
any required Skill Builder call, and the resulting commit complete before the
next task is processed. Later proposals in the same iteration can therefore
observe earlier commits, making evolution order-dependent. No iteration-$t$
Executor can observe these commits because all rollouts in that sweep have
already completed under $S_t^{\mathrm{Full}}$. After all task-level decisions
are processed, the updated live library is exported as
$S_{t+1}^{\mathrm{Full}}$ for the next sweep.

\section{Experiments}
\label{sec:experiments}

\subsection{Experimental Setup and Comparison Protocol}
\label{sec:experimental-setup}

\paragraph{Environment and tasks.}
We evaluate realistic Linux desktop operations drawn from OSWorld
\citep{xie2024osworld}, using framework version
\texttt{v0.1.16-934-g8f41f80a} and the
\texttt{happysixd/osworld-docker:latest} container. We use four application
domains---GIMP, VLC, LibreOffice Writer, and Thunderbird---with the fixed task
sets and iteration horizons in \cref{tab:domains}. Every iteration repeats the
same domain task set in the same order. We report one independently executed
longitudinal run for each condition--domain pair. For condition
$c\in\{\mathrm{Full},\mathrm{Empty}\}$, the official evaluator returns a
task-level score $s_{d,t,i}^{c}\in[0,1]$, potentially including partial
credit. We report the domain-level mean evaluator score on a percentage scale:
\begin{equation}
    \operatorname{Score}_{d,t}^{c} = \frac{100}{N_d}
    \sum_{i=1}^{N_d}s_{d,t,i}^{c}.
    \label{eq:mean-score}
\end{equation}

\begin{table}[ht]
    \centering
    \caption{Fixed task sets, matched iteration horizons, and external OSWorld
    reference scores. References contextualize the operating range and
    are not matched baselines.}
    \label{tab:domains}
    \small
    \setlength{\tabcolsep}{7pt}
    \begin{tabular}{lrrr}
        \toprule
        Application & Tasks $N_d$ & Iterations & OSWorld ref. (\%) \\
        \midrule
        GIMP         & 26 & 35 & 76 \\
        VLC          & 17 & 20 & 49 \\
        Writer       & 23 & 20 & 69 \\
        Thunderbird  & 15 & 40 & 80 \\
        \bottomrule
    \end{tabular}
\end{table}

\paragraph{Agent and evolution protocol.}
The Executor is EvoCUA-32B \citep{xue2026evocua}, and MVP
\citep{zhang2025mvp} provides screenshot--instruction grounding to absolute
coordinates. Kimi K2.5 \citep{kimi2026k25} serves as the Extractor in both
conditions and as the Proposer and Skill Builder in Full; all model parameters
remain fixed. Full and Empty use the same Executor--Grounding stack, task sets,
task order, iteration horizons, Executor prompt template, catalog interface,
and \texttt{get\_skill} tool; the catalog content is determined by each
condition's library state. Both start empty. Full executes iterations 0--4 with
empty snapshots, then exposes the first non-empty snapshot at iteration 5 and
follows the serial evolution protocol in \cref{sec:skill-evolution}. Empty runs
the same Extractor but disables the Proposer and Skill Builder, leaving its
catalog empty and clamping the library to $\varnothing$ throughout. Because the
conditions are independently executed, their empty-library warm-up trajectories
need not coincide.

\FloatBarrier
\subsection{Longitudinal Performance Comparison}
\label{sec:evolution-dynamics}

\begin{figure}[!ht]
    \centering
    \includegraphics[width=\linewidth]{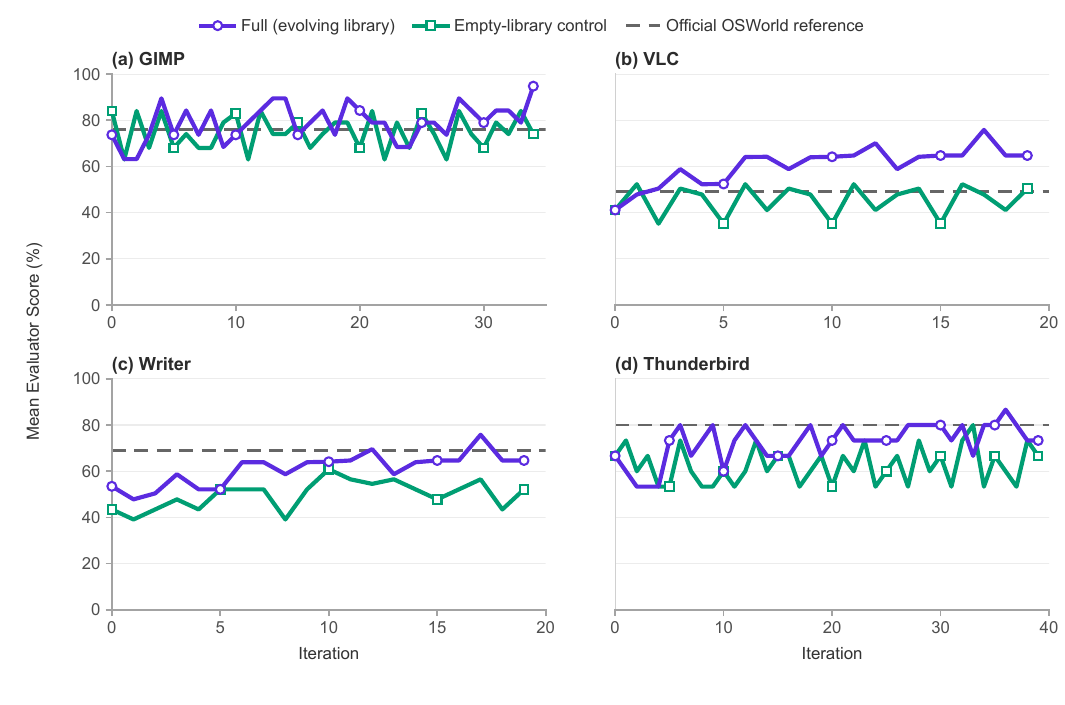}
    \caption{Longitudinal performance across four GUI application domains.
    Purple curves show the Full evolving-library system, for which the first
    non-empty state is available at iteration 5; teal curves show the matched
    Empty-Library Control. The independently executed conditions use the same
    fixed Executor--Grounding stack, task sets, and iteration horizons. Gray
    dashed lines are external OSWorld references, not matched baselines.}
    \label{fig:dynamics}
\end{figure}

\Cref{fig:dynamics} reports unsmoothed trajectories of the domain-level mean
evaluator score, while \cref{tab:full-empty} summarizes post-warm-up means
alongside pre-period differences. During the skill-enabled period ($t\geq5$),
Full has a higher mean
score than Empty Control in all four observed domain runs, with differences
ranging from 5.7 to 18.6 percentage points. VLC shows the largest difference
and a relatively stable post-warm-up separation. Writer and Thunderbird exhibit positive but more
variable post-warm-up margins, whereas GIMP has the smallest mean difference
and repeated crossings. The trajectories therefore show domain-dependent,
non-monotonic differences rather than uniform improvement at every iteration.

The independently executed conditions are already separated during the
empty-library warm-up. The Full-minus-Empty pre-period difference is negative
for GIMP ($-4.0$ points) and Thunderbird ($-6.7$), then reverses after the
library becomes available. VLC's positive difference widens from $+4.7$ to
$+18.6$ points. By contrast, Writer already differs by $+9.1$ points while both
libraries are empty, so most of its $+11.8$-point post-warm-up separation cannot
be uniquely associated with skill availability. With one longitudinal run per
condition--domain pair, these results provide a descriptive comparison of the
integrated evolving-library system, not a significance test or an isolated
causal estimate of skill revision.

\begin{table}[ht]
    \centering
    \caption{Post-warm-up mean evaluator scores and pre-period differences.
    Post means use $t\geq5$; pre differences use $t=0$--$4$. Differences are
    Full minus Empty Control in percentage points and are computed before
    rounding the displayed means.}
    \label{tab:full-empty}
    \small
    \setlength{\tabcolsep}{4.5pt}
    \begin{tabular}{lrrrr}
        \toprule
        Domain & Full post (\%) & Empty post (\%) & Post diff. (pp) & Pre diff. (pp) \\
        \midrule
        GIMP        & 80.2 & 74.4 & +5.7  & $-4.0$ \\
        VLC         & 64.0 & 45.4 & +18.6 & $+4.7$ \\
        Writer      & 63.9 & 52.0 & +11.8 & $+9.1$ \\
        Thunderbird & 74.7 & 62.3 & +12.4 & $-6.7$ \\
        \bottomrule
    \end{tabular}
\end{table}
\FloatBarrier

\subsection{Skill Retrieval and Revision Dynamics in GIMP}
\label{sec:gimp-skill-analysis}

GIMP provides the most detailed skill-level view of how the library evolves.
In the same 35-iteration run used in \cref{fig:dynamics}, the final library
contains 27 skills, and 82.4\% of valid rollouts invoke \texttt{get\_skill} at
least once. The library is therefore not merely accumulated as an offline
artifact; its procedures are routinely exposed to the Executor during
interaction. This statistic measures retrieval occurrence, not whether the
Executor follows the returned procedure or whether retrieval improves the
outcome.

Provenance further shows that retrieval extends beyond the tasks that produced
the skills. Among recorded skill calls, 56.7\% retrieve a skill whose origin is
the current task, whereas 43.3\% retrieve a skill created by another task. This
substantial cross-origin share provides direct evidence that the library is
consulted as shared procedural memory rather than only as task-specific
storage. It does not, however, establish causal transfer: telemetry identifies
which skill is retrieved, but not how strongly it influences subsequent
actions.

Repeated revision also does not guarantee task recovery. The
background-transparency task is the origin of three related skills:
\texttt{gimp-select-uniform-background},
\texttt{gimp-remove-uniform-background}, and
\texttt{gimp-add-alpha-channel}. Within this family,
\texttt{gimp-add-alpha-channel} undergoes repeated accepted edits, yet the task
succeeds in only 2 of 35 iterations (5.7\%). This pattern
constitutes \emph{revision churn}: failure evidence continues to induce library
changes without a corresponding improvement in the originating task. The case
demonstrates active but ineffective evolution and is consistent with a
bottleneck in the fixed Executor--Grounding stack, although it does not by
itself identify the failing component.

\section{Discussion and Limitations}
\label{sec:discussion}

The longitudinal comparison characterizes persistent skill evolution as useful
but conditional. Although Full has a higher post-warm-up mean evaluator score
in all four observed domains, the magnitude and temporal pattern of the
difference remain domain-dependent. In particular, Writer already exhibits a
substantial positive offset during the empty-library warm-up. These runs
therefore show that the integrated evolving-library system can add measurable
value to a fixed computer-use stack, but they do not establish that every raw
post-warm-up difference is caused by skill availability.

Provenance exposes both the shared nature and the limits of the resulting
memory. The substantial cross-origin retrieval share shows that stored
procedures are consulted beyond their tasks of origin, but retrieval alone does
not establish that the Executor follows or benefits from a skill. The
background-transparency case sharpens this distinction: repeated accepted edits
coexist with success in only 2 of 35 iterations. Because skills supply textual
procedural guidance while GUI actions still pass through the fixed
Executor--Grounding stack, revision cannot necessarily repair an underlying
execution or localization failure. This case is consistent with such a
bottleneck, but it does not identify whether the limiting component is skill
selection, action generation, or grounding.

Several limitations constrain broader interpretation. Each condition--domain
pair contains one independently executed run, leaving stochastic variation
unestimated and warm-up imbalance as a confound. Repeated fixed task sets
measure within-set adaptation rather than unseen-task transfer, and detailed
provenance currently covers only GIMP. External OSWorld scores are contextual,
while cross-origin retrieval remains observational. Serial evolution is also
order-dependent and lacks automatic consolidation or rollback. Immediate
priorities are replicated runs, held-out tasks, matched component ablations,
and direct measurement of skill adherence.

Beyond these gaps, two directions can extend the Skill Evolution paradigm.
First, it should be evaluated with agents that generate native computer-use
actions without an additional Grounding module, testing whether its value
generalizes beyond the current execution interface. Second, the evolution
policy itself can be made learnable. Revision churn shows that plausible
accepted mutations need not improve future utility; reinforcement-learning-based
optimization of the Proposer could instead reward downstream improvement and
penalize ineffective revisions, moving from prompted evidence-guided revision
toward learned long-horizon adaptation.

\section{Conclusion}
\label{sec:conclusion}

We presented an online skill-evolution framework that converts interaction
trajectories and evaluator feedback into persistent, versioned procedural
memory without changing model parameters. Iteration-frozen snapshots separate
execution from evidence-guided skill creation and revision, while provenance
records make the resulting lifecycle auditable. Across four realistic Linux
application domains drawn from OSWorld, Full records a higher post-warm-up mean
evaluator score than the configuration-matched Empty condition in every
observed run, although the magnitude and temporal pattern vary and warm-up
offsets limit causal interpretation. In GIMP, cross-origin retrieval shows that
skills are consulted beyond their tasks of origin, while revision churn shows
that repeated accepted edits can fail to recover the originating task.
Together, these findings position evolving skill libraries as shared,
auditable adaptation layers whose benefits remain domain-dependent rather than
guaranteed. Future work should test the paradigm with native action generation
without an additional Grounding module and use reinforcement learning to
optimize library-mutation policies for downstream utility.

\bibliography{references}
\bibliographystyle{tmlr}

\end{document}